\documentclass[10pt,twocolumn,letterpaper]{article}

\usepackage{wacv}              

\definecolor{wacvblue}{rgb}{0.21,0.49,0.74}
\usepackage[pagebackref,breaklinks,colorlinks,allcolors=wacvblue]{hyperref}
\usepackage{multirow, multicol}
\usepackage{epstopdf}
\usepackage{graphicx}
\usepackage{algorithm}
\usepackage[noend]{algpseudocode}
\usepackage{amsmath}
\usepackage{bbm}
\usepackage[symbol]{footmisc}

\def\wacvPaperID{2280} 
\def\confName{WACV}
\def\confYear{2026}

\title{MANTA: Physics-Informed Generalized Underwater Object Tracking}

\author{Suhas Srinath$^{1,*}$  \quad Hemang Jamadagni$^{2,*}$  \quad Aditya Chandrasekar$^{1,\dag}$ \quad Prathosh A P$^1$\\
$^1$Indian Institute of Science \quad $^2$ National Institute of Technology Karnataka\\
}

\begin{document}
\maketitle


\begin{abstract}
Underwater object tracking is challenging due to wavelength-dependent attenuation and scattering, which severely distort appearance across depths and water conditions. Existing trackers trained on terrestrial data fail to generalize to these physics-driven degradations. We present MANTA, a physics-informed framework integrating representation learning with tracking design for underwater scenarios. We propose a dual-positive contrastive learning strategy coupling temporal consistency with Beer--Lambert augmentations to yield features robust to both temporal and underwater distortions. We further introduce a multi-stage pipeline augmenting motion-based tracking with a physics-informed secondary association algorithm that integrates geometric consistency and appearance similarity for re-identification under occlusion and drift. To complement standard IoU metrics, we propose Center–Scale Consistency (CSC) and Geometric Alignment Score (GAS) to assess geometric fidelity. Experiments on four underwater benchmarks (WebUOT-1M, UOT32, UTB180, UWCOT220) show that MANTA achieves state-of-the-art performance, improving Success AUC by up to $6\%$, while ensuring stable long-term generalized underwater tracking and efficient runtime. Code available at \url{https://github.com/Kazedaa/MANTA}.
\end{abstract}



\section{Introduction}

Underwater object tracking (UOT) poses challenges distinct from terrestrial tracking due to severe image degradations from wavelength-dependent attenuation, scattering, and color absorption~\cite{dcp, survey2, survey3, survey4}. These factors drastically alter object appearance across depths and conditions, creating a significant domain gap that hinders a direct transfer from terrestrial trackers~\cite{generaluot1, generaluot2, generaluot3}. Similar to general tracking, UOT includes single-object tracking (SOT) and multi-object tracking (MOT), but progress has been driven largely by SOT~\cite{uwcot, utb180, uot100, webuot, uvot400} given the scarcity of MOT datasets~\cite{brackishmot}.
Most existing UOT approaches adapt terrestrial methods via preprocessing or fine-tuning~\cite{uostrack, webuot, uvot400}, yet such methods fail to capture physics-driven distortions inherent to underwater imaging. As a result, trackers struggle with wavelength-selective absorption, scattering, and depth-dependent contrast loss. This work focuses on advancing underwater SOT by explicitly modeling these domain-specific challenges.
\footnote[1]{Equal contribution. $^\dag$Work done while at Indian Institute of Science.}

\begin{figure}[!t]
    \centering
    \includegraphics[width=\columnwidth]{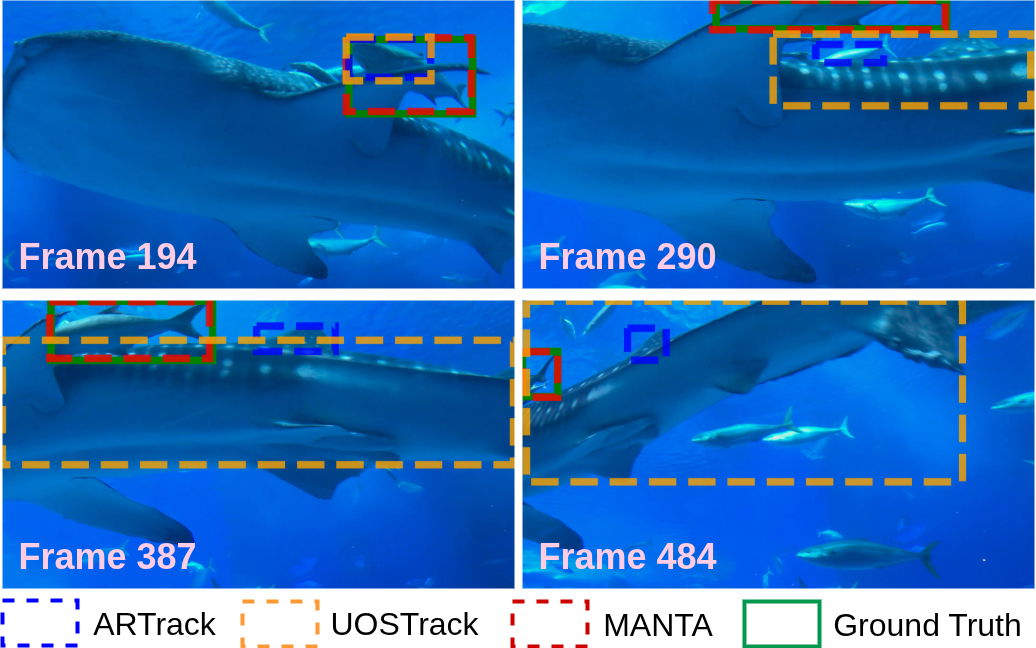}
    \caption{Tracking outputs from different methods on \texttt{Video\_0002} from the UTB180~\cite{utb180} dataset. MANTA (dashed red) reliably tracks the target correct object (solid green) across frames, maintaining identity even under occlusion or temporary disappearance.}
    \label{fig:teaser}
\end{figure}

Recent advances in self-supervised learning (SSL) offer a promising alternative, enabling the learning of powerful visual representations without reliance on labeled data~\cite{selfsupervisedsurvey}. In particular, contrastive learning has demonstrated strong capability in learning semantically meaningful and generalizable features by enforcing consistency across differently augmented views of the same instance~\cite{generalssl1, generalssl2, generalssl3}. Yet, most SSL pipelines employ domain-agnostic augmentations such as random cropping, color jittering, or geometric transformations. These transformations fail to capture underwater-specific distortions, limiting their applicability to UOT.

To address these challenges, we introduce a physics-informed self-supervised framework for underwater SOT.
Our dual-positive contrastive learning strategy combines temporal consistency with physics-based priors by constructing positive pairs from (1) temporally aligned object crops across frames and (2) Beer--Lambert~\cite{beer} augmented crops simulating depth-dependent attenuation. This enforces invariance to both temporal and underwater distortions, yielding domain-generalizable features for UOT.

Beyond representation learning, robust tracking must address occlusion, disappearance, and drift. Lightweight real-time trackers perform well terrestrially~\cite{ocsort, deepsort, bytetrack} but degrade underwater due to overlapping objects, domain gaps, and frequent identity loss~\cite{deepocsort}. We mitigate these issues by augmenting a motion-based primary tracker with a physics-informed secondary association algorithm that leverages visual cues for long-term re-identification, improving temporal coherence and stability under challenging conditions.

Finally, we introduce two metrics for finer-grained evaluation: Center--Scale Consistency (CSC), evaluating both positional accuracy and bounding-box scale, and Geometric Alignment Score (GAS), applying Gaussian penalties to jointly capture center and scale misalignment.
In summary, we present \textbf{MANTA} (\textbf{MA}ri\textbf{N}e object \textbf{T}racking \textbf{A}lgorithm), a physics-informed self-supervised framework for underwater SOT that integrates robust representation learning into a physics-informed vision-based secondary association algorithm for efficient and generalizable tracking. Our contributions are summarized as follows:
\begin{itemize}
    \item We propose a dual-positive contrastive learning framework that combines temporal consistency with Beer--Lambert augmentations, enabling features independent of both temporal variations and underwater distortions.
    
    \item We design a multi-stage tracker that augments motion cues with a physics-informed secondary association using geometric consistency and appearance similarity. Cascaded strategies of ID reuse, geometric scoring, and localized appearance search ensure stable long-term trajectories under occlusion, clutter, and low visibility.
    
    \item We propose two new evaluation metrics: Center-Scale Consistency (CSC), requiring accurate center and scale prediction, and Geometric Alignment Score (GAS), applying Gaussian penalties for misalignment to provide finer tracker evaluation insights beyond IoU.
    
    \item Extensive experiments on four underwater benchmarks show consistent state-of-the-art performance, with improvements in Success AUC, geometric alignment, and long-term stability. 
\end{itemize}


\section{Related Work}

\subsection{Underwater Object Tracking}

Underwater object tracking is critical for marine robotics, autonomous vehicles, and biology~\cite{generaluot1, uotapplication1, uotapplication2, uotapplication3}. Early adaptations of terrestrial trackers using preprocessing such as histogram equalization and color correction~\cite{preprocess1}), struggle with underwater distortions. Later designs introduced underwater-specific correlation filters~\cite{survey2} and adaptive feature selection~\cite{uotapplication1}, but these hand-crafted methods fail to generalize across environments. Deep learning–based trackers~\cite{artrack, transt, seqtrack, siamfc, siamrpn} improved performance, yet their reliance on terrestrial pretraining (e.g., ImageNet~\cite{imagenet}, COCO~\cite{coco}, GOT-10k~\cite{got10k}) created domain gaps. Classical real-time trackers~\cite{bytetrack, ocsort, deepsort} are efficient but often fail under occlusion or when re-identifying lost objects. Our work addresses these gaps by learning domain-robust features via physics-informed self-supervised learning, coupled with a secondary vision-guided association algorithm that complements motion-based tracking for improved re-identification.

\begin{figure*}[!t]
    \centering
    \includegraphics[width=\linewidth]{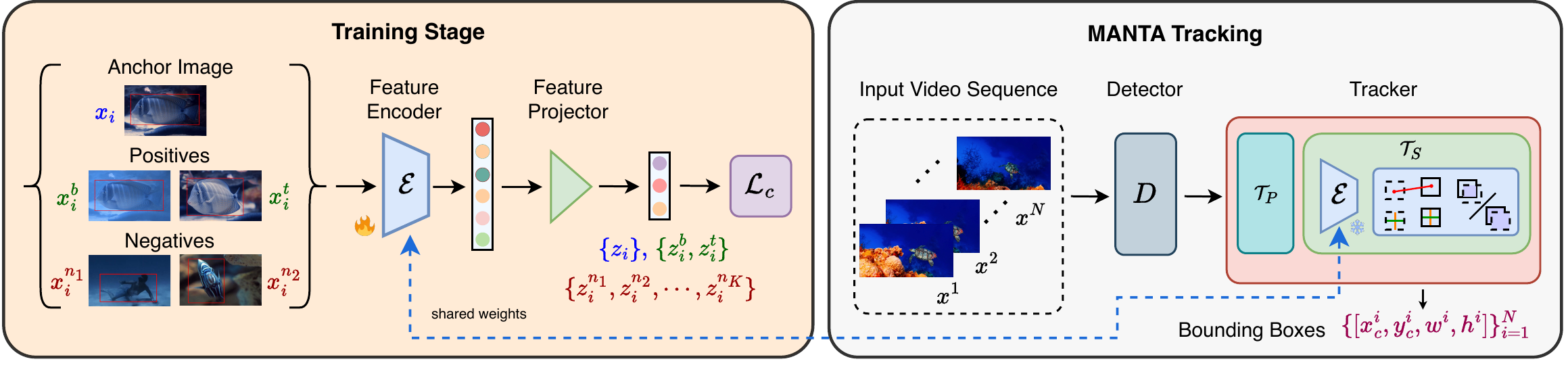}
    \caption{Overview of MANTA. The self-supervised encoder $\mathcal{E}$ is trained via contrastive learning with Beer–Lambert augmentations $x_i^b$ and temporal augmentations $x_i^t$. Following detections from $D$ and primary tracking with $\mathcal{T}_P$, embeddings produced by $\mathcal{E}$ are used for vision-guided secondary association $\mathcal{T}_S$, which refines trajectories by matching IoU, scales, and centers, yielding accurate predictions for an input video sequence.}
    \label{fig:main_block}
\end{figure*}

\subsection{Self-Supervised Learning for Visual Tracking}

Self-supervised learning has enabled robust representation learning without manual annotations~\cite{generalssl1, generalssl2, generalssl3}, and has recently been explored for visual tracking~\cite{uotssl1}. Contrastive learning, in particular, has shown strong potential by constructing positive pairs from temporal sequences or spatial augmentations to enforce independence~\cite{uotssl2, uotssl3}. For instance, \citeauthor{contrastive}~\cite{contrastive} showed that contrastive training on tracking sequences can outperform ImageNet pretraining, while other works leverage object crops across frames~\cite{crops} or optical flow correspondences~\cite{featmatch}.  
However, most self-supervised tracking methods rely solely on temporal consistency and generic augmentations, overlooking domain-specific information. In contrast, our approach integrates temporal contrastive learning with physics-informed augmentations (depth-based), yielding a dual-supervision framework that captures both global object identity and is robust to underwater distortions.

\subsection{Physics-Informed Underwater Vision}

The integration of physical principles into machine learning has gained significant attention across scientific domains~\cite{physicsml1, physicsml2}. In computer vision, physics-based models have been applied to image restoration~\cite{imagerest}, depth estimation, and domain adaptation~\cite{domainadapt}. Central to underwater imaging is the Beer–Lambert law~\cite{beer, beer2}, which models wavelength-dependent light attenuation and provides a principled description of image degradation caused by absorption, scattering, and backscattering~\cite{seathru}. Prior works have primarily exploited this physics for enhancement and restoration~\cite{phishnet, undive}, including Dark Channel Prior adaptations~\cite{dcp} and GAN-based underwater-to-terrestrial mappings~\cite{gan}.  
In contrast, our work employs these physical models not for image correction, but as augmentations within a self-supervised framework. By embedding Beer–Lambert–based transformations into temporal contrastive learning, we explicitly learn features independent of underwater distortions while preserving semantic object identity. This strategy is different from generic augmentation pipelines and enables the learning of domain-independent features that generalize effectively to underwater object tracking.

\section{Methodology}
\label{sec:methodology}

Our approach addresses the fundamental challenge of underwater object tracking through a physics-informed self-supervised learning framework. The key insight is that underwater imaging follows predictable physical principles, which we can exploit to learn robust representations that generalize across diverse underwater conditions. Our complete pipeline consists of three interconnected components: (1) a dual-positive contrastive learning framework that learns domain-invariant features, (2) physics-based augmentations that simulate realistic underwater distortions, and (3) a multi-stage tracking pipeline that leverages these learned representations for robust object association. Figure~\ref{fig:main_block} provides an overview of our complete methodology.

\subsection{Physics-Informed Representation Learning}

The core of our approach lies in learning representations that are invariant to physical distortions in underwater environments. We achieve this through a dual-positive contrastive learning strategy that enforces consistency across both temporal variations and physics-based augmentations.

\subsubsection{Dual-Positive Contrastive Framework}

Traditional contrastive learning methods for tracking~\cite{uotssl1, uotssl2, uotssl3} primarily exploit temporal correspondence or generic augmentations such as color jitter, motion blur, and saturation. Our framework extends this approach by introducing physics-informed augmentations alongside temporal consistency, creating a dual-supervision mechanism that captures both object identity persistence and robustness to underwater distortions.

For each target object bounding box crop $\mathbf{x}_i$ at time $t$, we construct two types of positive pairs:

\begin{itemize}
    \item \textbf{Temporal Positives}: Object crops $\mathbf{x}_{t+\Delta t}$ from subsequent frames containing the same object, enforcing temporal consistency of object identity. These crops are denoted as $x_i^t$ with corresponding features $z_i^t$.  
    \item \textbf{Physics Positives}: Beer–Lambert~\cite{beer,beer2} augmented versions of the current crop, simulating depth-dependent underwater distortions. These are denoted as $x_i^b$ with features $z_i^b$.  
\end{itemize}

Negatives are defined as object crops from other sequences in the batch, denoted by $\{x_i^{n_j}\}_{j=1}^K$ with corresponding features $\{z_i^{n_j}\}_{j=1}^K$. Our contrastive learning objective, based on InfoNCE~\cite{infonce} and supervised contrastive loss~\cite{supcon}, encourages similarity between the anchor representation $z_i$ and both temporal and physics positives, while pushing away negatives:

\begin{equation}
    \mathcal{L}_c = \frac{1}{N_b} \sum_{i=1}^{N_b} \mathcal{L}_i
\end{equation}
where the per-sample loss is defined as  
\begin{equation}
    \nonumber
    \mathcal{L}_i = -\log 
    \frac{
        \exp\!\left(f_\theta(z_i) \cdot f_\theta(z_i^t) / \tau\right) 
        + 
        \exp\!\left(f_\theta(z_i) \cdot f_\theta(z_i^b) / \tau\right)
    }{
        \sum_{k=1}^K \exp\!\left(f_\theta(z_i) \cdot f_\theta(z_i^{n_k}) / \tau\right)
    }
\end{equation}
with $\tau$ denoting the temperature parameter. This formulation drives the encoder $\mathcal{E}$ to produce features that align representations of the same object across time and varying physical conditions, while maintaining separation from different objects.

\subsubsection{Beer-Lambert Physics Augmentations}

To generate realistic physics-informed augmentations, we leverage the Beer-Lambert law, which governs light attenuation in underwater environments~\cite{beer2}. This physical model allows us to simulate the depth-dependent appearance variations that objects naturally experience underwater.
Given a training image $\mathbf{I}$ and its corresponding depth map $\mathbf{D}$ (obtained using UDepth~\cite{udepth}, an efficient off-the-shelf monocular depth estimator), we compute the Beer-Lambert augmented image as:
\begin{equation}
\mathbf{I}^{BL}(\mathbf{p}) = \mathbf{I}(\mathbf{p}) \cdot T(\mathbf{p}) + \mathbf{B} \cdot (1 - T(\mathbf{p}))
\end{equation}
where $\mathbf{p}$ denotes pixel coordinates, $T(\mathbf{p}) = e^{-\beta \mathbf{D}(\mathbf{p})}$ represents the transmission map, $\beta$ is the attenuation coefficient, and $\mathbf{B}$ represents the background water color.
During training, we vary $\beta \in [0.1, 0.5]$ within each batch to simulate diverse depth-dependent attenuation, while keeping a fixed blue–green illumination $\mathbf{B}$. Lower $\beta$ values yield minimal distortion, whereas higher values risk obscuring objects, ensuring a balance between augmentation diversity and object visibility. By varying $\beta$ across training batches, the encoder learns to extract features that are invariant to different attenuation levels, enabling generalization across diverse underwater environments without overfitting to any specific water type or depth range. Importantly, Beer–Lambert augmentation serves purely as a training strategy to expose the model to realistic distortions; hence, precise depth accuracy is unnecessary since the objective is invariance across attenuation patterns. Depth maps are used only during training, with inference relying solely on RGB inputs.

\subsection{Multi-Stage Tracking Pipeline}

Our tracking framework integrates the learned physics-informed representations into a three-stage pipeline that combines the efficiency of motion-based tracking with the robustness of vision-guided association.

\subsubsection{Primary Motion-Based Tracking}

We begin with OC-SORT~\cite{ocsort} as our primary tracker, which provides efficient motion-based association through optical flow consistency and temporal smoothness. While OC-SORT is effective for short-term tracking, it is susceptible to identity switches in cluttered underwater scenes and track loss during occlusions.

\subsubsection{Vision-Guided Secondary Association}

To address the limitations of motion-only tracking, we introduce a vision-guided secondary association module that leverages our physics-informed features to maintain track continuity. This module operates through a sequential matching strategy designed to recover from track loss while maintaining computational efficiency.
Starting from the ground-truth anchor in frame one, the algorithm maintains tracking continuity through the following steps:
\begin{enumerate}
    \item \textbf{Initialization:} In the first frame, we associate the ground-truth anchor with the detection that maximizes a composite geometric score. For two bounding boxes $b_1$ and $b_2$, we define:
    \begin{align}
    \text{dist}(b_1,b_2) &= 1 - \frac{\|c(b_1) - c(b_2)\|_2}{\max(\text{diag}(b_1), \text{diag}(b_2))} \\
    \text{scale}(b_1,b_2) &= \frac{\min(\text{area}(b_1), \text{area}(b_2))}{\max(\text{area}(b_1), \text{area}(b_2))}
    \end{align}
    The composite score combines both spatial and scale consistency as:
    \begin{align}
    \nonumber \text{score}(b_1,b_2) &= w_{dist}\cdot \text{dist}(b_1,b_2) + w_{scale}\cdot \text{scale}(b_1,b_2)
    \end{align}
    
    \item \textbf{History-based Track Reuse:} For subsequent frames, we first check if any recently seen track IDs reappear with geometric consistency.
    
    \item \textbf{Active Track Retention:} If the current target ID remains active and shows consistency, we retain its bounding box.
    
    \item \textbf{Score-based Re-acquisition:} When the target ID is lost, we re-select candidates using the composite score relative to the previous track state.
    
    \item \textbf{Physics-Informed Local Search:} As a final fallback, we perform local search around the last known position, using our physics-informed encoder $\mathcal{E}$ to compute cosine similarity between candidate crops and the target representation.
\end{enumerate}
The similarity function for geometric and semantic consistency is defined as:
\begin{equation}
\scalebox{0.9}{$
\begin{aligned}
  \text{are\_similar}(b_1,b_2)
    = \mathbbm{1}\!\Big[
       &\text{size\_ok}(b_1,b_2)
       \land
       \bigl(\text{IoU}(b_1,b_2) > 0\bigr) \\[0.8ex]
       &\land
       \bigl(\cos\bigl(\mathcal{E}(b_1), \mathcal{E}(b_2)\bigr) \ge \theta_{cos}\bigr)
     \Big]
\end{aligned}
$}
\end{equation}
where $\wedge$ denotes logical conjunction (AND), and $\mathcal{E}(b_1)$ and $\mathcal{E}(b_2)$ denote the representations of the object crop with bounding boxes $b_1$ and $b_2$, respectively. The $\text{size\_ok}(\cdot,\cdot)$ function and IoU are defined as
\begin{equation}
\scalebox{0.925}{$
  \vcenter{
    \hbox{$
      \begin{gathered}
        \text{size\_ok}(b_1,b_2)
          = \mathbbm{1}\!\left[
                \frac{\max(w_1,w_2)}{\min(w_1,w_2)} \le 2
                \;\land\;
                \frac{\max(h_1,h_2)}{\min(h_1,h_2)} \le 2
            \right] \\[1ex]
        \text{IoU}(b_1,b_2)
          = \frac{|b_1 \cap b_2|}{|b_1 \cup b_2|}
      \end{gathered}$
    }
  }
$}
\end{equation}
where $\mathbbm{1}(\cdot)$ denotes the indicator function, $w_1$ and $w_2$ are the widths of the two bounding boxes, and $h_1$ and $h_2$ are the heights.
The complete algorithm is summarized in Algorithm~\ref{algo}.

\begin{algorithm}[t]
\caption{Vision-Guided Secondary Association}
\label{algo}
\begin{algorithmic}[1]
\Function{Secondary\_Association}{Primary\_Tracks, anchor}
    \State $history.push(-1)$ \Comment{LRU Stack}
    \For{each $frame$ in sequence}
        \State $dets=detections[$frame$]$
        \If{$history.peek(d.id)==-1$}
            \State $d^*=\arg\max_{d \in dets}\text{score}(anchor, d.bbox)$
            \State $curr\_bbox=anchor$
            \State $history.push(d^*.id)$
        \Else 
            \For{each $h \in$ history and $d \in dets$}
                \If{$d.id==h$}
                    \If{$\text{are\_similar}(curr\_bbox,d.bbox)$}
                        \State $history.push(h)$
                        \State $curr\_bbox=d.bbox$
                        \State \textbf{break both loops}
                    \EndIf
                \EndIf
            \EndFor
            \For{$d \in dets$}
                \If{$d.id==history.peek(d.id)$}
                    \If{$\text{are\_similar}(curr\_bbox,d.bbox)$}
                        \State $curr\_bbox=d.bbox$
                        \State \textbf{break loop}
                    \EndIf
                \EndIf
            \EndFor
            
            \State $d^*=\arg\max_{d \in dets}\text{score}(anchor, d.bbox)$
            \State $curr\_bbox=d^*.bbox$
            \State $history.push(d^*.id)$
            \State \textbf{break loop}
            \State $c^*=\arg\max_{c \in \text{crops}} \; \cos\big(\mathcal{E}(c), \mathcal{E}(curr\_bbox)\big)$
            \If{$\cos\big(\mathcal{E}(c^*), \mathcal{E}(curr\_bbox)\big) \geq \theta_{cos}$}
                \State $curr\_bbox \gets c^*.bbox$
            \EndIf
        \EndIf
        \State append $curr\_bbox$ to predictions
    \EndFor
    \State \Return predictions
\EndFunction
\end{algorithmic}
\end{algorithm}

\subsubsection{End-to-End Integration}

Our complete system integrates these components through a three-stage pipeline: (1) \textbf{Object Detection} using fine-tuned RF-DETR~\cite{rfdetr} on curated underwater datasets, (2) \textbf{Motion-Guided Association} via OC-SORT for efficient short-term tracking, and (3) \textbf{Vision-Guided Re-association} using our physics-informed features to recover lost targets and maintain long-term consistency.
This design leverages the complementary strengths of motion-based efficiency and vision-based robustness, while the physics-informed representations provide the domain-specific invariances necessary for reliable underwater tracking.


\section{Experiments and Results}

\begin{table*}[htbp]
\centering
\resizebox{0.9\textwidth}{!}{
\begin{tabular}{clccccccc}
\toprule
Dataset & Method & Suc. AUC & Suc.@0.5 & Pre. AUC & Pre.@20px & mIoU & mGAS & mCSC@0.2 \\
\midrule
\multirow{10}{*}{\rotatebox{90}{\textbf{UOT32~\cite{uot32}}}} & SiamFC & 0.4226 & 0.4823 & 0.4499 & 0.4673 & 0.4955 & 0.5778 & 0.1893 \\
& STARK & 0.6153 & 0.7498 & 0.6432 & 0.7057 & 0.6906 & 0.7671 & 0.5627 \\
& SiamRPN++ & 0.6204 & 0.7636 & 0.6452 & 0.6943 & 0.6782 & 0.7811 & 0.5732 \\
& TransT & 0.6693 & 0.8068 & 0.7097 & 0.7661 & \textcolor{blue}{0.7333} & 0.8239 & 0.6096 \\
& SeqTrack & 0.6581 & 0.8027 & \textcolor{blue}{0.7217} & \textcolor{blue}{0.7932} & 0.7116 & 0.8373 & 0.6075 \\
& OSTrack & 0.6640 & 0.8132 & 0.7074 & 0.7685 & 0.7097 & 0.8288 & 0.6195 \\
& OKTrack & 0.6307 & 0.7744 & 0.6727 & 0.7448 & 0.6811 & 0.7973 & 0.5852 \\
& ARTrack & 0.6612 & 0.8058 & 0.7181 & 0.7769 & 0.7235 & \textcolor{blue}{0.8381} & 0.5984 \\
& UOSTrack & \textcolor{blue}{0.6700} & \textcolor{blue}{0.8300} & 0.7096 & 0.7795 & 0.7285 & 0.8337 & \textcolor{blue}{0.6285} \\
\midrule
& \textbf{MANTA} & \textcolor{purple}{0.7306} & \textcolor{purple}{0.8848} & \textcolor{purple}{0.7628} & \textcolor{purple}{0.8419} & \textcolor{purple}{0.7632} & \textcolor{purple}{0.8829} & \textcolor{purple}{0.7438} \\
\bottomrule
\vspace{-2.5mm}\\
\multirow{10}{*}{\rotatebox{90}{\textbf{UTB180~\cite{utb180}}}} & SiamFC & 0.3571 & 0.3870 & 0.2470 & 0.2333 & 0.5101 & 0.4836 & 0.2056 \\
& STARK & 0.5439 & 0.6224 & 0.4358 & 0.4547 & 0.7172 & 0.6347 & 0.5066 \\
& SiamRPN++ & 0.5478 & 0.6546 & 0.4243 & 0.4233 & 0.6775 & 0.6856 & 0.4755 \\
& TransT & 0.5852 & 0.6685 & 0.4836 & 0.5086 & 0.7450 & 0.6876 & 0.5458 \\
& SeqTrack & 0.6184 & 0.7099 & 0.5228 & 0.5528 & 0.7830 & 0.7160 & 0.6046 \\
& OSTrack & 0.6454 & 0.7380 & 0.5399 & 0.5691 & 0.7611 & 0.7499 & 0.6121 \\
& OKTrack & 0.7014 & \textcolor{blue}{0.7999} & 0.6100 & \textcolor{blue}{0.6623} & 0.7860 & \textcolor{blue}{0.8123} & \textcolor{blue}{0.7010} \\
& ARTrack & \textcolor{blue}{0.7085} & 0.7980 & \textcolor{blue}{0.6215} & 0.6548 & \textcolor{blue}{0.8021} & 0.8099 & 0.6706 \\
& UOSTrack & 0.6608 & 0.7581 & 0.5588 & 0.5894 & 0.7703 & 0.7689 & 0.6370 \\
\midrule
& \textbf{MANTA} & \textcolor{purple}{0.7328} & \textcolor{purple}{0.8268} & \textcolor{purple}{0.6698} & \textcolor{purple}{0.7383} & \textcolor{purple}{0.8351} & \textcolor{purple}{0.8312} & \textcolor{purple}{0.7746} \\
\bottomrule
\vspace{-2.5mm}\\
\multirow{10}{*}{\rotatebox{90}{\textbf{UW-COT-220~\cite{uwcot}}}} & SiamFC & 0.2830 & 0.3007 & 0.1937 & 0.1743 & 0.4606 & 0.3922 & 0.1580 \\
& STARK & 0.4707 & 0.5345 & 0.3682 & 0.3700 & 0.5968 & 0.5736 & 0.3979 \\
& SiamRPN++ & 0.4268 & 0.4994 & 0.3410 & 0.3363 & 0.6193 & 0.5458 & 0.3313 \\
& TransT & 0.4998 & 0.5729 & 0.4128 & 0.4184 & 0.6686 & 0.6224 & 0.4246 \\
& SeqTrack & 0.5638 & 0.6507 & 0.4934 & 0.5121 & 0.6806 & 0.6849 & 0.5012 \\
& OSTrack & 0.5504 & 0.6288 & 0.4677 & 0.4846 & 0.6555 & 0.6638 & 0.4973 \\
& OKTrack & 0.5976 & 0.6935 & 0.5162 & 0.5391 & 0.6684 & 0.7205 & 0.5378 \\
& ARTrack & \textcolor{purple}{0.6075} & \textcolor{blue}{0.6964} & \textcolor{purple}{0.5385} & \textcolor{purple}{0.5616} & \textcolor{blue}{0.6966} & \textcolor{blue}{0.7283} & \textcolor{blue}{0.5404} \\
& UOSTrack & 0.5747 & 0.6629 & 0.4926 & 0.5116 & 0.6639 & 0.6909 & 0.5204 \\
\midrule
& \textbf{MANTA} & \textcolor{blue}{0.6037} & \textcolor{purple}{0.7036} & \textcolor{blue}{0.5255} & \textcolor{blue}{0.5480} & \textcolor{purple}{0.7631} & \textcolor{purple}{0.7376} & \textcolor{purple}{0.5936} \\
\bottomrule
\vspace{-2.5mm}\\
\multirow{10}{*}{\rotatebox{90}{\textbf{WebUOT-1M~\cite{webuot}}}} & SiamFC & 0.3224 & 0.3487 & 0.2397 & 0.2282 & 0.4728 & 0.4432 & 0.1724 \\
& STARK & 0.5341 & 0.6189 & 0.4507 & 0.4735 & 0.6848 & 0.6467 & 0.4909 \\
& SiamRPN++ & 0.4961 & 0.5928 & 0.4073 & 0.4137 & 0.6255 & 0.6315 & 0.4203 \\
& TransT & 0.5466 & 0.6352 & 0.4754 & 0.4971 & 0.6874 & 0.6721 & 0.4925 \\
& SeqTrack & 0.5647 & 0.6506 & 0.5005 & 0.5244 & 0.6993 & 0.6839 & 0.5239 \\
& OSTrack & 0.5919 & 0.6850 & 0.5221 & 0.5477 & 0.6887 & 0.7188 & 0.5491 \\
& OKTrack & 0.6294 & 0.7218 & 0.5579 & 0.5943 & 0.7012 & 0.7527 & \textcolor{blue}{0.6031} \\
& ARTrack & \textcolor{blue}{0.6393} & \textcolor{blue}{0.7287} & \textcolor{blue}{0.5793} & \textcolor{blue}{0.6099} & \textcolor{blue}{0.7155} & \textcolor{purple}{0.7689} & 0.5846 \\
& UOSTrack & 0.6109 & 0.7067 & 0.5346 & 0.5618 & 0.6719 & 0.7377 & 0.5699 \\
\midrule
& \textbf{MANTA} & \textcolor{purple}{0.6417} & \textcolor{purple}{0.7319} & \textcolor{purple}{0.5843} & \textcolor{purple}{0.6342} & \textcolor{purple}{0.7564} & \textcolor{blue}{0.7538} & \textcolor{purple}{0.6594} \\
\bottomrule
\vspace{-2.5mm}\\
\end{tabular}
}
\caption{Performance comparison of tracking methods across four datasets. Best and second best results are highlighted in \textcolor{purple}{purple} and \textcolor{blue}{blue} respectively. For all the metrics, higher value indicates better performance.}
\label{tab:main_table}
\end{table*}

\begin{table*}[!ht]
\centering
\setlength{\tabcolsep}{0.5pt}
\begin{tabular}{ccccc} 
\includegraphics[width=0.2\textwidth]{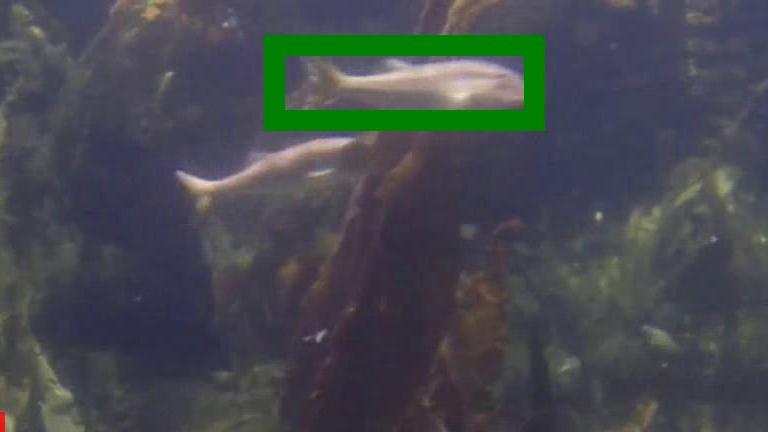}     &    \includegraphics[width=0.2\textwidth]{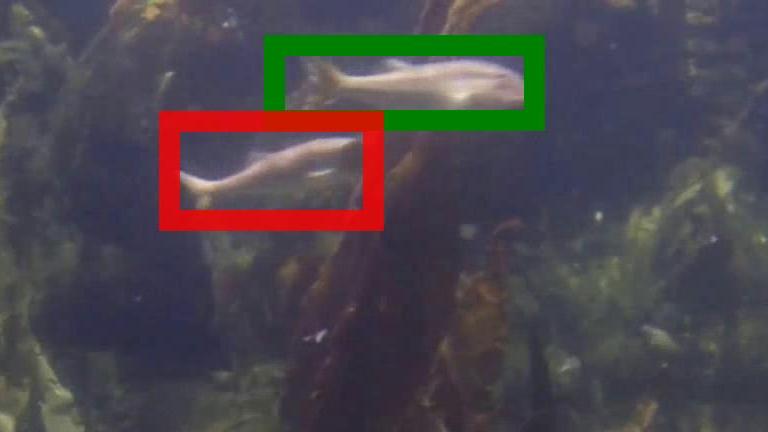} & \includegraphics[width=0.2\textwidth]{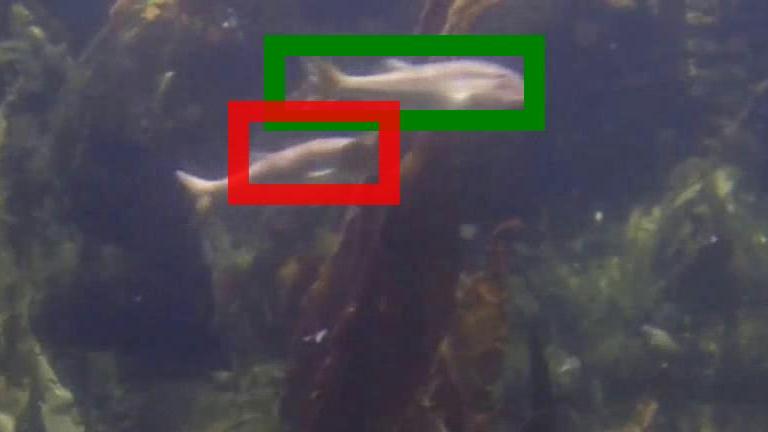}     & 
\includegraphics[width=0.2\textwidth]{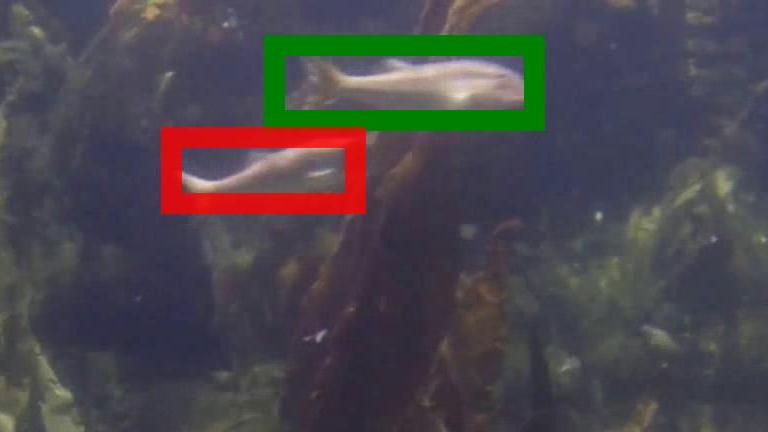}  &   \includegraphics[width=0.2\textwidth]{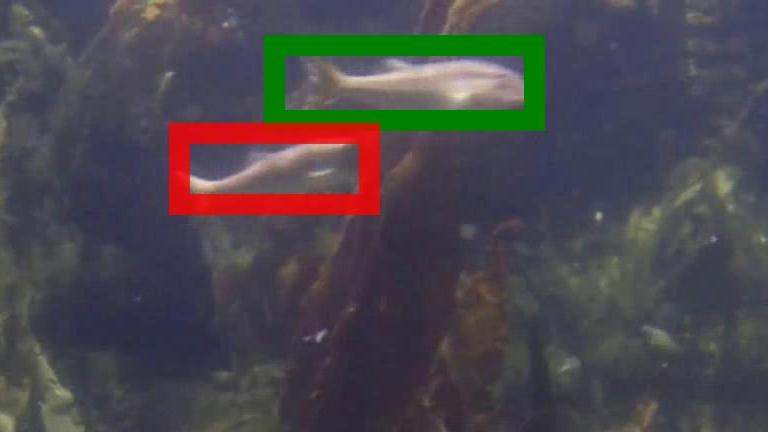}  \\
SiamFC & SiamRPN++ & STARK & TransT & OSTrack \vspace{1mm}\\
\includegraphics[width=0.2\textwidth]{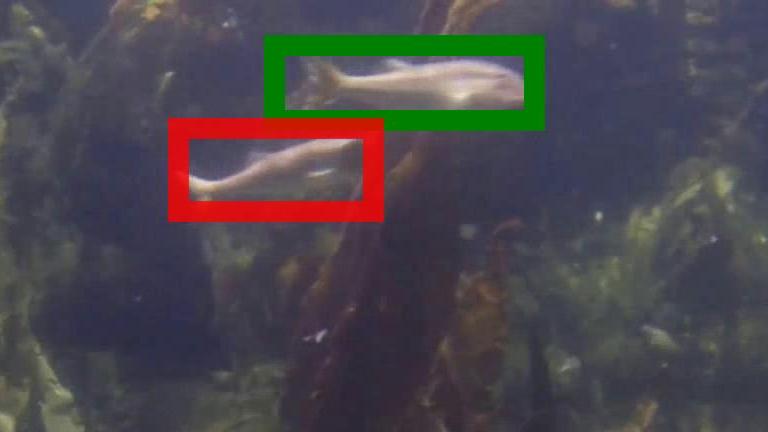}     &    \includegraphics[width=0.2\textwidth]{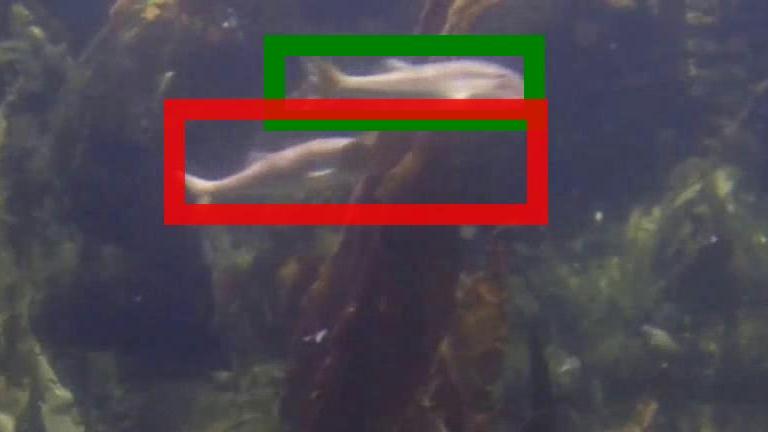}     & 
\includegraphics[width=0.2\textwidth]{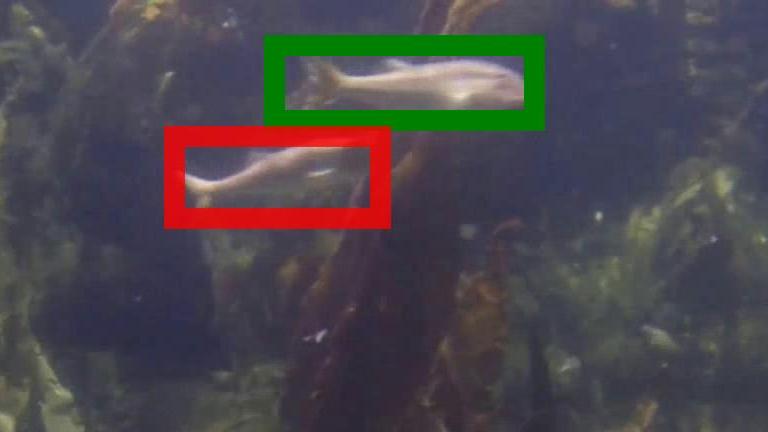}  &   \includegraphics[width=0.2\textwidth]{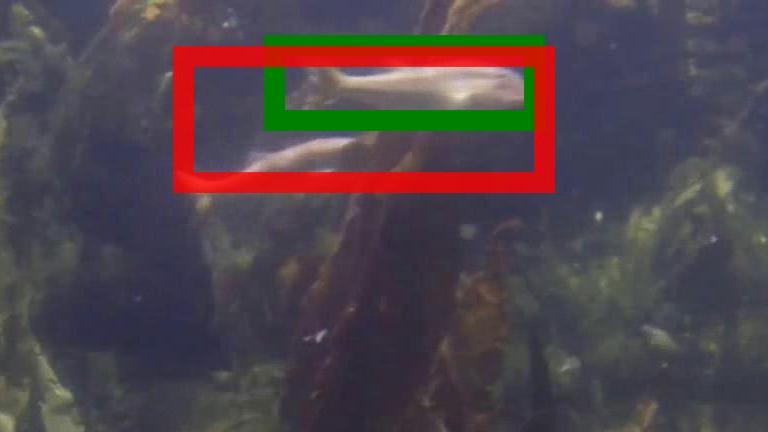}     &  \includegraphics[width=0.2\textwidth]{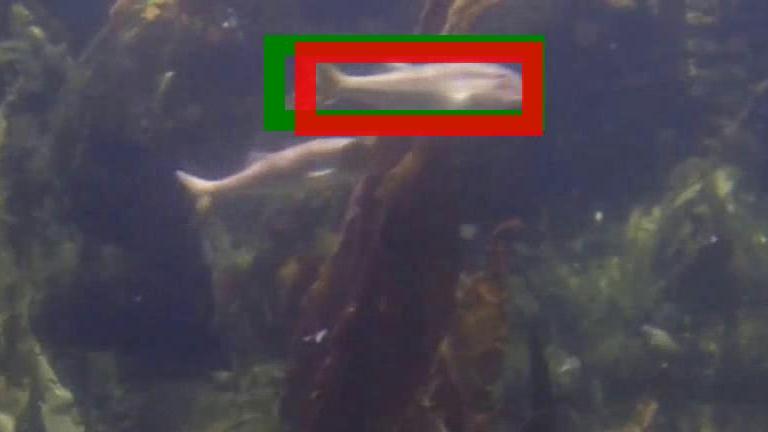} \\
OKTrack & SeqTrack & ARTrack & UOSTrack & MANTA (Ours) \\ 
\end{tabular}
\captionof{figure}{Comparison of tracking methods on frame 261 of the \texttt{WebUOT-1M\_Test\_000388} sequence from the WebUOT-1M~\cite{webuot} dataset. Tracker outputs are shown in \textcolor{red}{red}, and ground-truth boxes in \textcolor{teal}{green}. Unlike other methods that either drift, miss the target, or track incorrect objects, MANTA consistently identifies and follows the correct object.}
\label{fig:comp_images}
\end{table*}

\subsection{Experimental Setup and Training Details}

All experiments were done on a 48GB NVidia RTX A6000 GPU. The detector was trained using a batch size of $B=16$, and the physics encoder with a batch size of $N_b=32$.

\textbf{Detector Training:} We fine-tune RF-DETR~\cite{rfdetr} on a curated underwater dataset combining SUIM~\cite{suim}, UIIS10k~\cite{uiis}, frames from UVOT~\cite{uvot400}, WebUOT-1M~\cite{webuot} (the official train split), UOT100~\cite{uot100} (excluding UOT32), and Brackish MOT~\cite{brackishmot}, covering diverse underwater scenes. Missing object class labels were manually annotated, and copy-paste augmentation was used to balance rare classes. The detector was trained for 50 epochs with the standard RF-DETR losses.

\vspace{1mm}
\textbf{Physics Encoder Training:} We train the encoder with contrastive learning on frames sampled from WebUOT-1M (train set), UVOT400, and Brackish MOT. Temporal positives $x_i^t$ are generated by sampling frames within four steps of the anchor, while Beer–Lambert augmentations $x_i^b$ use a fixed illumination map $\mathbf{B} = [0.6, 0.8, 0.9]$ (RGB) to model typical blue-green water conditions in SOT datasets. Attenuation strength $\beta$ is varied in $[0.1, 0.5]$ to simulate depth-dependent distortions. A ResNet-18~\cite{resnet} backbone with a 128-dimensional projection head is trained using InfoNCE~\cite{infonce} loss and optimized with AdamW~\cite{adam} at a learning rate of $10^{-5}$ for 100 epochs.

\vspace{1mm}
\textbf{Tracker-specific parameters:} For primary association, we use default OC-SORT~\cite{ocsort} parameters. The secondary association uses a sliding window stride factor of 0.5 (half the bounding box dimensions), a search expansion factor of 1.5, $w_{dist} = w_{size} = 1$ and $\theta_{cos} = 0.9$ for the matching of the appearance (lower thresholds led to noisy matches, hindering the re-identification process). These thresholds were selected through validation on a held-out subset of the training data to balance matching precision and recall in the re-identification process.

\subsection{Evaluation Datasets}

\textbf{WebUOT-1M}~\cite{webuot}: a large-scale benchmark with over 1M frames covering diverse marine organisms, vehicles, and objects under varying water conditions and depths. We report results only on the test set for fair comparison.
\textbf{UOT32}~\cite{uot32}: 32 challenging sequences featuring fish, divers, and underwater equipment across environments with varying visibility.
\textbf{UTB180}~\cite{utb180}: 180 high-quality sequences emphasizing difficult cases such as poor visibility, color distortion, and occlusion by marine life.
\textbf{UW-COT-220}~\cite{uwcot}: 220 sequences focusing on camouflaged object tracking, spanning diverse lighting, clarity, and object categories.

\subsection{Evaluation Metrics}

We evaluate our method using both standard tracking metrics and two novel metrics specifically designed to capture fine-grained tracking performance.

\textbf{Standard Metrics:} We report Success AUC and Precision AUC, which are widely used in visual tracking literature~\cite{generaluot1}. Success AUC measures the area under the curve of success rates at different overlap thresholds, while Precision AUC measures the area under the curve of precision rates at different center location error thresholds. We also report Success@0.5 and Precision@20px as threshold-specific metrics, along with mean IoU (mIoU).

\textbf{Center-Scale Consistency (CSC):} While conventional precision measures only the alignment of bounding-box centers, it ignores the accuracy of the predicted box dimensions. To jointly evaluate both factors, we introduce Center-Scale Consistency (CSC). For a predicted bounding box $b_p = (x_p, y_p, w_p, h_p)$ and the ground truth $b_g = (x_g, y_g, w_g, h_g)$, we first compute the normalized center error:
\begin{equation}
e_c = \frac{\|c_p - c_g\|_2}{\sqrt{w_g^2 + h_g^2}}, \quad c_p = \left(x_p + \frac{w_p}{2}, y_p + \frac{h_p}{2}\right),
\end{equation}
where $c_g = \left(x_g + \frac{w_g}{2}, y_g + \frac{h_g}{2}\right)$. We then define the relative scale errors as:
\begin{equation}
e_w = \frac{|w_p - w_g|}{w_g}, \quad e_h = \frac{|h_p - h_g|}{h_g}.
\end{equation}
CSC is computed as the fraction of frames where both the center and scale errors fall below user-specified thresholds $(\tau_c, \tau_s)$:
\begin{equation}
\text{CSC}(\tau_c, \tau_s) = \frac{1}{N} \sum_{i=1}^{N} \mathbbm{1}\left(e_c^i < \tau_c \land e_w^i < \tau_s \land e_h^i < \tau_s\right),
\end{equation}
where $\mathbbm{1}(\cdot)$ is the indicator function and the operator $\wedge$ denotes logical conjunction (AND). Intuitively, CSC measures how consistently the tracker predicts both the correct location and scale of the target. 

\textbf{Geometric Alignment Score (GAS):} To provide a continuous measure of geometric fidelity, we propose the Geometric Alignment Score (GAS), which softly penalizes both misalignment and scale mismatch. Given the normalized center error $e_c$ and scale error:
\begin{equation}
e_s = \frac{(w_p - w_g)^2 + (h_p - h_g)^2}{w_g^2 + h_g^2},
\end{equation}
we define:
\begin{equation}
\text{GAS} = \exp\left(-\frac{e_c^2}{\sigma_c^2}\right) \cdot \exp\left(-\frac{e_s}{\sigma_s^2}\right),
\end{equation}
where $\sigma_c$ and $\sigma_s$ are tolerance parameters controlling sensitivity. GAS takes values in $(0, 1]$, with higher scores indicating better geometric alignment between prediction and ground truth.

In our experiments, we set $\tau_c = \tau_s = 0.2$ for CSC to require tight geometric alignment, and $\sigma_c = \sigma_s = 0.5$ for GAS to provide moderate tolerance for small misalignments.

\subsection{Quantitative Results}

\textbf{Comparing Methods:} We compare MANTA with nine state-of-the-art trackers, including Siamese networks (SiamFC~\cite{siamfc}, SiamRPN++~\cite{siamrpn}), transformer-based models (STARK~\cite{stark}, TransT~\cite{transt}, SeqTrack~\cite{seqtrack}, ARTrack~\cite{artrack}, OSTrack~\cite{ostrack}), and underwater-specific approaches (UOSTrack~\cite{uostrack}, OKTrack~\cite{webuot}). 

Results across four benchmark datasets (Table~\ref{tab:main_table}) show MANTA consistently outperforms prior methods. On the UOT32~\cite{uot32} dataset, it achieves 0.7306 Success AUC and 0.8829 Success@0.5, surpassing UOSTrack by 5–6\%. Our metrics reveal even larger gains: +11.5\% CSC@0.2 and +4.2\% mGAS. On UTB180~\cite{utb180}, MANTA improves Success AUC (0.7328 vs. 0.7085 for ARTrack) and excels in alignment (0.8312 mGAS) and scale consistency (0.7746 CSC@0.2). For UW-COT220, it achieves the best mIoU (0.7631) and Success@0.5 (0.7036). On the large-scale WebUOT-1M, MANTA delivers consistent gains, including the top CSC@0.2 (0.6594). These improvements highlight the benefits of physics-informed contrastive learning and the vision-based re-identification, which enforces physical and track consistency, yielding good performance under low visibility, clutter, and scale variations.

\subsection{Qualitative Results}

Qualitative analysis demonstrates that MANTA maintains robust tracking across diverse underwater scenarios, including severe color distortions, low-visibility conditions, complex marine environments, and the presence of multiple distractor objects. The physics-informed representations enable consistent target identity preservation across frames, even under challenging conditions such as occlusion, low visibility, and appearance variations.
As shown in Fig.~\ref{fig:comp_images}, MANTA accurately identifies and follows the target fish, while competing methods such as STARK, TransT, OSTrack, and ARTrack drift towards incorrect objects, UOSTrack tracks multiple objects simultaneously, and SiamFC fails to detect the target altogether. In contrast, MANTA leverages an efficient re-identification mechanism to recover from occlusions and maintain the correct trajectory despite challenging underwater conditions. More detailed analyses and discussions are provided in the supplementary.

\subsection{Ablation Studies}

We validate the individual contributions of our proposed components through systematic ablations on the UOT32 dataset (Table~\ref{tab:ablation}).  
The baseline tracker achieves 0.7000 Success AUC and 0.7280 Precision AUC.  
Adding ImageNet-pretrained ResNet50 features offers only minor gains, underscoring the limited value of generic representations for UOT.  
Temporal contrastive learning improves performance to 0.7259 / 0.7576, while Beer–Lambert augmentations perform slightly better (0.7286 / 0.7596), showing the benefit of physics-informed independence.  
Combining both yields the best results: 0.7306 Success AUC, 0.7628 Precision AUC, and 0.7438 mCSC@0.2, with consistent improvements in mGAS.  
These results demonstrate that temporal and physics-based augmentations are complementary, and together enable the most robust underwater tracking.

\begin{table}[t]
\centering
\resizebox{\columnwidth}{!}{%
\begin{tabular}{l|ccccc}
\hline
Method & Succ. AUC & Prec. AUC & mIoU & mGAS & mCSC@0.2 \\
\hline
None & 0.7000 & 0.7280 & 0.7463 & 0.8560 & 0.6928 \\
ResNet18 & 0.7046 & 0.7311 & 0.7653 & 0.8512 & 0.7221 \\
Physics - T & 0.7259 & 0.7576 & 0.7648 & 0.8767 & 0.7401 \\
Physics - BL & 0.7286 & 0.7596 & 0.7646 & 0.8802 & 0.7433 \\
Physics - T + BL & \textbf{0.7306} & \textbf{0.7628} & \textbf{0.7653} & \textbf{0.8829} & \textbf{0.7438} \\
\hline
\end{tabular}%
}
\caption{Ablation study results on UOT32 dataset. We analyze the impact of encoder training in our MANTA framework. T and BL correspond to temporal and Beer-Lambert augmentations, respectively. Physics - T corresponds to the use of only temporal augmentaions.}
\label{tab:ablation}
\end{table}

\subsection{Runtime and Complexity Analysis}

\begin{table}[t]
\centering
\resizebox{0.5\columnwidth}{!}{%
\begin{tabular}{lcc}
\hline
Method    & FPS & Parameters \\ \hline
SiamFC    & 30  & 2.3M       \\
SiamRPN++ & 35  & N/A         \\
ARTrack   & 29  & 86M        \\
OKTrack   & 52  & 92.1M      \\
OSTrack   & 23  & 86M        \\
SeqTrack  & 31  & 86M        \\
STARK     & 17  & 87M        \\
TransT    & 40  & 23M        \\
UOSTrack  & 56  & N/A         \\
MANTA     & 37  & 45.4M      \\ \hline
\end{tabular}}%
\caption{Runtime and complexity analysis of tracking methods evaluated on the UOT32~\cite{uot32} dataset (N/A - not available).}
\label{tab:runtime}
\end{table}

\begin{figure}
    \centering
    \includegraphics[width=\columnwidth]{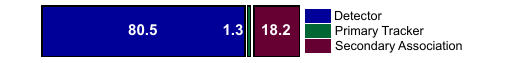}
    \caption{Runtime breakdown percentages of total runtime.}
    \label{fig:runtime_analysis}
\end{figure}

Table~\ref{tab:runtime} reports runtime (FPS) and parameter counts on UOT32~\cite{uot32}. SiamFC is extremely lightweight (2.3M) but lacks capacity, while TransT offers a good trade-off (40 FPS, 23M). Heavy transformer trackers such as ARTrack, OSTrack, SeqTrack, and STARK (86–87M) run at 17–31 FPS, and OKTrack scales to 92.1M with 52 FPS, showing strong optimization but high memory cost.
MANTA achieves 37 FPS with 45.4M parameters—about half the size of heavy transformer trackers while maintaining competitive speed. Compared to UOSTrack (56 FPS, complexity not reported), MANTA offers a clearer balance of efficiency and robustness. As shown in Fig.~\ref{fig:runtime_analysis}, most computation lies in detection (80.5\%), while tracking (1.3\%) and re-association (18.2\%) are lightweight. Overall, MANTA provides an efficient speed–accuracy trade-off suited for real-time underwater tracking. \\ 

\noindent \textbf{Limitations:} Despite strong performance, MANTA has limitations. It relies on monocular depth estimation, which may be unreliable in texture-poor or repetitive scenes, though precise depth is unnecessary since we target invariance across attenuation patterns rather than precise physics modeling. The secondary association module uses fixed thresholds, limiting adaptability to varying conditions; adaptive thresholding could improve robustness. Finally, while effective for SOT, extending to MOT remains challenging due to the complexity of maintaining multiple identities under severe appearance distortions.

\section{Conclusion}

We introduced MANTA, a physics-informed framework for underwater object tracking that integrates Beer–Lambert–based augmentations into a dual-positive contrastive learning strategy. This design enables representations robust to temporal variations and underwater degradations. Integrated into a novel motion and vision-guided association tracker, MANTA yields consistent state-of-the-art performance across four underwater SOT benchmarks. In addition, our CSC and GAS metrics provide a more precise evaluation of geometric fidelity. Overall, MANTA demonstrates that embedding physical priors into self-supervised learning is a powerful direction for building robust and generalizable tracking systems in challenging real-world domains.

\subsection*{Acknowledgements}

Suhas Srinath acknowledges support from the Ministry of Education (MoE), India.


{
    \small
    \bibliographystyle{ieeenat_fullname}
    \bibliography{camera_ready}
}

\end{document}